\documentclass[journal,twoside,web]{IEEEtran}
\usepackage{cite}
\usepackage{amsmath,amssymb,amsfonts}
\usepackage{algorithmic}
\usepackage{algorithm}
\usepackage{graphicx}
\usepackage{textcomp}
\usepackage{xcolor}
\usepackage{booktabs} 
\usepackage{multirow}
\usepackage{url}     
\usepackage{subfigure} 
\usepackage{tikz}
\usetikzlibrary{shapes.geometric, arrows.meta, positioning, calc, decorations.pathreplacing}
\usepackage{enumitem}
\graphicspath{{./figures/}}
\begin{document}

\title{Causal-Transformer with Adaptive Mutation-Locking for Early Prediction of Acute Kidney Injury}

\author{Weizhi Nie, Haolin Chen, Huifang Hao$^\ast$ Yuting Su, Keliang Xie, Bo Yang
  \thanks{Weizhi Nie, Haolin Chen and Yuting Su are with the School of Electrical and Information Engineering, Tianjin University.}
  \thanks{$^\ast$Huifang Hao is with the Department of Nephrology, TEDA Hospital, Tianjin University. (Corresponding Author: 280591008@qq.com)}
  \thanks{Keliang Xie is with the Tianjin Medical University General Hospital.}
  \thanks{Bo Yang is with the Department of Nephrology, First Teaching Hospital of Tianjin University of Traditional Chinese Medicine.}
}

\maketitle
\begin{abstract}
  Accurate early prediction of Acute Kidney Injury (AKI) is critical for timely clinical intervention. However, existing deep learning models struggle with irregularly sampled data and suffer from the opaque ``black-box'' nature of sequential architectures, strictly limiting clinical trust.
  To address these challenges, we propose CT-Former, integrating continuous-time modeling with a Causal-Transformer. To handle data irregularity without biased artificial imputation, our framework utilizes a continuous-time state evolution mechanism to naturally track patient temporal trajectories. To resolve the black-box problem, our Causal-Attention module abandons uninterpretable hidden state aggregation. Instead, it generates a directed structural causal matrix to identify and trace the exact historical onset of severe physiological shocks. By establishing clear causal pathways between historical anomalies and current risk predictions, CT-Former provides native clinical interpretability. Training follows a decoupled two-stage protocol to optimize the causal-fusion process independently. Extensive experiments on the MIMIC-IV cohort (N=18,419) demonstrate that CT-Former significantly outperforms state-of-the-art baselines. The results confirm that our explicitly transparent architecture offers an accurate and trustworthy tool for clinical decision-making.
\end{abstract}


\section{Introduction}
\label{sec:introduction}
\IEEEPARstart{A}{cute} kidney injury (AKI) is a complex organ dysfunction syndrome characterized by a rapid decline in renal function. It is a frequent complication in hospitalized patients, particularly in intensive care units (ICUs), and is associated with exceptionally high morbidity and mortality rates. According to global health data, this condition affects millions of lives annually with a continuously escalating trend\cite{ronco2019acute}. The clinical management of AKI is critically time-sensitive. However, current global standards largely rely on delayed physiological biomarkers such as serum creatinine, meaning that overt clinical diagnosis often lags behind the optimal therapeutic window. Given that early identification of high-risk patients can significantly improve prognosis by enabling preemptive interventions---such as optimized fluid management and the avoidance of nephrotoxic agents---precise data-driven early prediction prior to clinical manifestation has become imperative to afford clinicians crucial preparatory time.\cite{tomasev2019}.

\begin{figure}[t]
  \centering
  \includegraphics[width=0.99\linewidth]{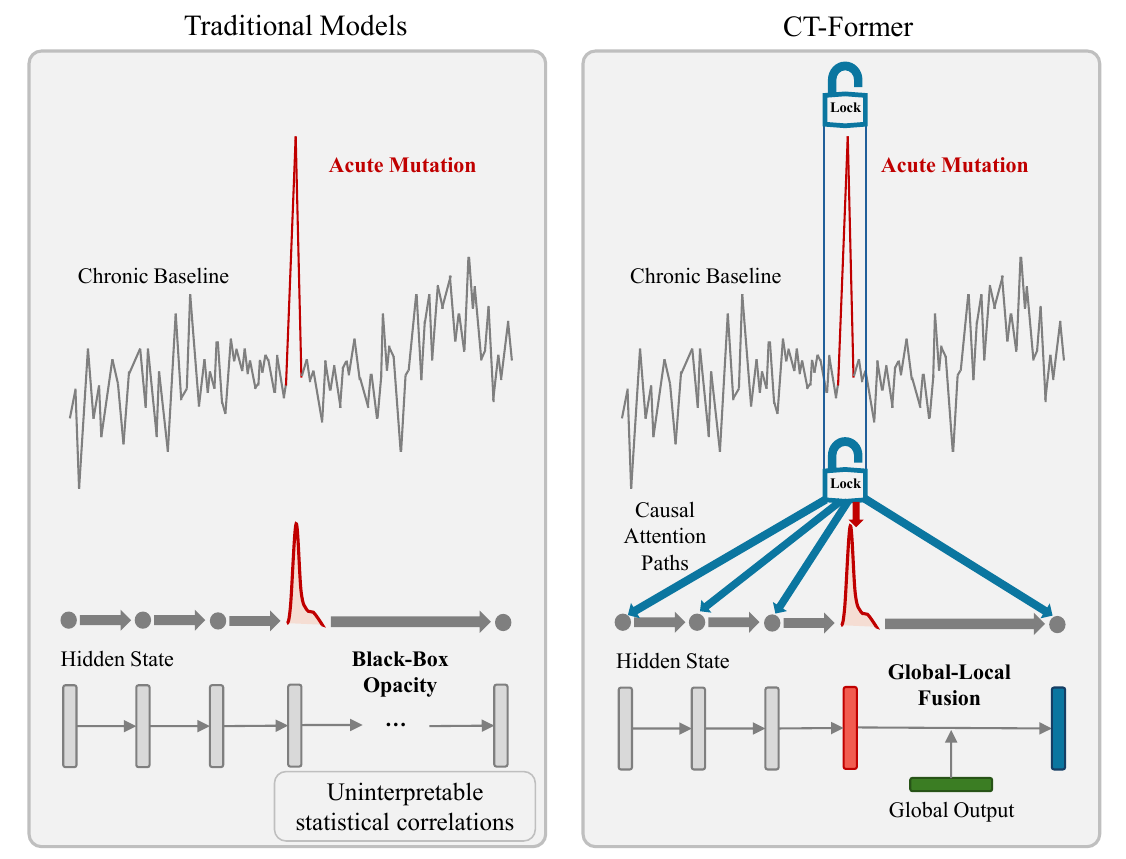}
  \caption{Comparison of sequential modeling approaches. Left: Traditional recurrent models suffer from black-box opacity and rely on uninterpretable statistical correlations. Right: Our CT-Former utilizes explicit causal attention to generate a structural causal matrix, transparently linking historical physiological mutations to current risk predictions.}
  \label{fig:intro}
\end{figure}

Electronic health records (EHR) exhibit severe sparsity and highly irregular sampling intervals \cite{tipirneni2022self,t1}. Traditional discrete-time architectures struggle to align asynchronous data, often requiring biased data imputation that distorts underlying biological processes. More critically, as illustrated in Fig. \ref{fig:intro} (Left), the ``black-box'' nature of sequential deep learning creates a significant barrier to clinical adoption. Existing models operate on uninterpretable state compounding and statistical correlations \cite{bai2018empirical}. They fail to construct verifiable pathways linking historical physiological events to subsequent predictions. In clinical practice, predictive alarms lacking pathological transparency are frequently distrusted. Consequently, two main research questions remain:
\begin{itemize}
  \item How to effectively model latent physiological trajectories from irregularly sampled EHR data?
  \item How to extract intrinsic causal relationships to overcome the black-box dilemma and provide transparent clinical interpretability?
\end{itemize}

To address these challenges, we discard opaque sequence modeling paradigms. First, we adopt a continuous-time state evolution mechanism to handle arbitrary observation gaps dynamically, eliminating arbitrary discrete alignments \cite{hasani2021liquid}. Second, we introduce a causal transformer architecture to build a structurally transparent reasoning process. Rather than relying on statistically aggregated hidden states, our model directly yields a directed causal matrix to identify precise historical moments of physiological shocks \cite{zibo}. This extracts clear causal pathways between historical inputs and clinical outputs, explicitly endowing the model with native transparency and reliable interpretability.

The main contributions of this paper are summarized as follows:
\begin{itemize}
  \item We propose CT-Former, which utilizes a continuous-time evolution mechanism to natively model irregular clinical intervals without biased data imputation.
  \item We develop a structural Causal-Attention module that resolves black-box opacity by generating a directed causal matrix to explicitly trace physiological triggers.
  \item Extensive experiments on the MIMIC-IV dataset demonstrate that CT-Former significantly outperforms established baselines, with independent TimeSHAP validation confirming its strong clinical interpretability.
\end{itemize}

\section{Related Work}
\label{sec:related_work}
\subsection{Clinical Diagnosis and Biomarkers of AKI}
Acute Kidney Injury (AKI) is a severe condition characterized by a sudden deterioration in renal function, which frequently occurs in critically ill patients admitted to the Intensive Care Unit (ICU). The clinical consensus for diagnosing AKI has evolved significantly over the past two decades. Initially, the ADQI workgroup introduced the RIFLE criteria to standardize the classification of renal dysfunction \cite{bellomo2004}. This was subsequently refined by the Acute Kidney Injury Network (AKIN) to capture less severe, yet clinically relevant, fluctuations in kidney health \cite{mehta2007}. Currently, the global standard is the KDIGO guideline, which identifies AKI based on absolute or relative elevations in serum creatinine (SCr) and sustained declines in urine output \cite{kdigo2012}.

Despite the widespread adoption of the KDIGO criteria, the reliance on SCr poses a fundamental challenge for timely intervention. SCr is widely recognized as a delayed physiological biomarker; its concentration in the blood typically remains within normal ranges until more than 50\% of the glomerular filtration rate has been compromised. By the time a clinical diagnosis is confirmed via SCr thresholds, patients have often missed the optimal therapeutic window for fluid resuscitation or nephrotoxin avoidance. To overcome this diagnostic lag, the medical informatics community has increasingly turned toward data-driven early warning systems designed to detect subtle, multivariate physiological shifts long before overt clinical symptoms emerge\cite{guohonglin}.

\subsection{Machine Learning for Early AKI Prediction}
Early predictive models largely treat electronic health records as static snapshots. For instance, tree-based models such as extreme gradient boosting\cite{chen2016xgboost} and random forest algorithms\cite{breiman2001random} have demonstrated efficacy in predicting acute kidney injury. Furthermore, studies confirm that such static machine learning models consistently outperform conventional clinical scoring methods. However, these algorithms fundamentally discard the chronological ordering of clinical events, failing to capture longitudinal health trajectories and dynamic physiological changes\cite{lidongchen}.

To capture sequential dependencies, researchers transitioned to discrete recurrent neural networks. Toma{\v{s}}ev \textit{et al.} demonstrated that recurrent architectures could successfully predict acute kidney injury up to forty eight hours in advance \cite{tomasev2019}. Nevertheless, standard recurrent models mandate fixed temporal intervals. This forces the use of imputation heuristics such as forward filling for sporadically sampled medical data. This distorts underlying biological signals and introduces synthetic noise \cite{che2018recurrent}. Furthermore, these purely sequential architectures operate as opaque statistical black boxes\cite{t2}. They fail to explicitly isolate the exact causal triggers of short term physiological mutations critical for early AKI detection.

To overcome imputation limitations, continuous-time methodologies integrated differential equations with recurrent architectures. Chen \textit{et al.} introduced neural ordinary differential equations to model hidden states continuously \cite{chen2018neural}. Building upon this, Rubanova \textit{et al.} developed ODE RNN to natively handle irregularly sampled sequences \cite{rubanova2019latent}. Becker \textit{et al.} proposed recurrent Kalman networks with elapsed time integration to incorporate temporal uncertainty \cite{becker2019recurrent}. While theoretically elegant for tracking continuous clinical baselines, these networks still rely on highly entangled mathematical associations. They lack the structural transparency to uncover the causal relationships behind sudden episodic spikes.

Concurrently, transformer architectures emerged to capture long range dependencies and preserve global contexts through self attention mechanisms \cite{vaswani2017attention}. Recognizing that standard discrete positional encodings struggle to natively model physiological decay, Chen \textit{et al.} proposed ContiFormer. This combines neural ordinary differential equations with continuous time attention \cite{chen2023contiformer}. However, architectures relying on numerical solvers like ContiFormer typically consume prohibitive amounts of training time and memory. This limits their suitability for resource constrained clinical settings.

Parallel to attention mechanisms, structural approaches such as transformable patching graph neural networks have been explored to capture dependencies in irregular multivariate time series \cite{patchgnn}. Despite these structural advancements, buffering and reprocessing long sequences incurs substantial computational overhead. Crucially, these models still lack a dedicated mechanism to eliminate the black box nature of their predictions. Consequently, the clinical paradigm is shifting toward interpretable and computationally efficient architectures. These models must track long term chronic trajectories while explicitly revealing the causal structure of sudden physiological crashes. We discuss this in the subsequent section.

\subsection{Continuous-Time Dynamics and Temporal Explainability}
To bypass the computational limitations of numerical solvers in differential equation networks, Hasani \textit{et al.} introduced liquid time-constant networks inspired by biological synapses \cite{hasani2021liquid}. Building upon this foundation, they subsequently developed closed-form continuous-time neural networks \cite{hasani2022}. By deriving a tightly bounded mathematical approximation, these closed-form networks completely eliminate numerical solvers while preserving continuous dynamics, enabling highly efficient modeling of irregularly sampled signals. Leveraging these advancements, our study proposes CT-Former, an architecture that intertwines the scalable continuous-time capabilities of closed-form networks with the global contextual reach of transformers.

Alongside architectural advancements, the clinical deployment of deep learning remains hindered by opaque decision-making. Traditional explainability methods—such as local agnostic models \cite{ribeiro2016why}, raw attention weights \cite{choi2016retain, jain2019attention}, and standard SHAP \cite{lundberg2017}—exhibit critical flaws for longitudinal records, as they inherently ignore temporal ordering, produce misleading narratives, or cause computationally prohibitive combinatorial explosions by flattening sequences. To overcome these limitations, we evaluate our CT-Former architecture leveraging TimeSHAP \cite{bento2021}, a temporal extension of the Shapley framework engineered for sequential models. By utilizing a temporal coalition pruning algorithm that aggregates distant historical events into a background state, TimeSHAP efficiently focuses on recent critical physiological fluctuations. This integration extracts multi-granular, clinically verifiable narratives across both temporal-event and feature-level attributions for early acute kidney injury prediction.

\section{Data Processing}
\label{sec:data_processing}
\subsection{Data Source and Cohort Selection}
This study utilized the MIMIC-IV database (v3.1), a large-scale, publicly available critical care dataset sourced from the electronic health records of the Beth Israel Deaconess Medical Center between 2008 and 2019\cite{johnson2020mimic}. Institutional Review Board approval was obtained with individual patient consent waived.

\begin{figure}[t]
  \centering
  \includegraphics[width=0.9\linewidth]{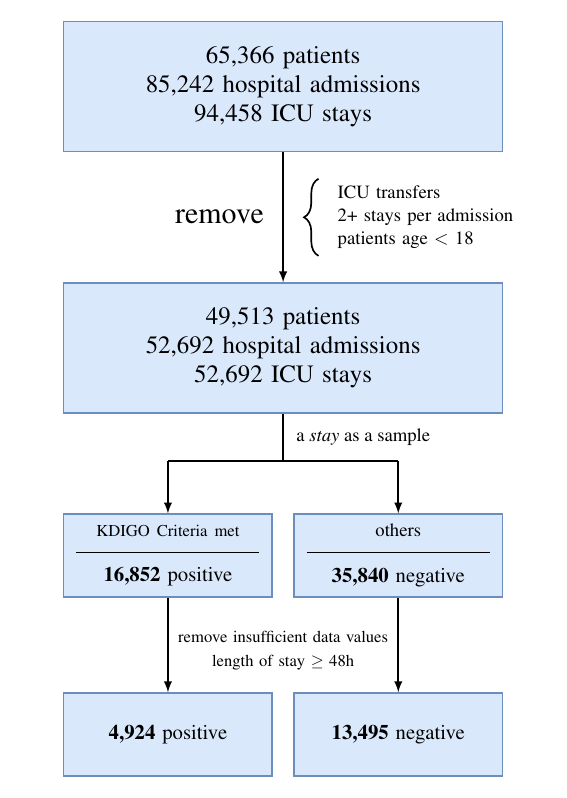}
  \caption{Flowchart for dataset processing and sample extraction. The first step removes ICU transfer records, multiple admissions, and patients under 18 years of age. In the second step, each ICU stay is treated as an independent sample; samples meeting KDIGO criteria are labeled as positive, while others are labeled as negative. The final step excludes samples with high data loss and retains only ICU stays with a duration of 48 hours or more.}
  \label{fig:cohort_flowchart}
\end{figure}
To ensure clinical validity and data quality, we applied a strict multi-stage filtering process (Fig. \ref{fig:cohort_flowchart}) to an initial retrieval of 85,242 ICU admissions. First, pediatric patients ($<$ 18 years) were excluded, as pediatric AKI involves distinct etiological mechanisms compared to adult physiology\cite{kaddourah2017epidemiology}. To maintain sample independence, we retained only the first ICU stay per subject, excluding cases with multiple admissions or intra-hospital transfers.

Subsequent filtering focused on observational sufficiency and baseline relevance. We excluded ICU stays shorter than 48 hours to ensure an adequate observation window for capturing preceding physiological trends. Patients missing $>40\%$ of core vital signs were removed to prevent computational bias. Finally, patients with a prior history of End-Stage Renal Disease (ESRD) or chronic dialysis were excluded, as their abnormal baseline renal function obscures acute injury risks\cite{hoste2015epidemiology}. After filtering, the final study cohort comprised 18,419 unique patients.

\subsection{Label Definition and Feature Engineering}

The ground truth for AKI was strictly defined following KDIGO guidelines \cite{kdigo2012}: a serum creatinine (SCr) increase $\ge$ 0.3 mg/dL within 48 hours, SCr $\ge$ 1.5$\times$ baseline within 7 days, or urine output $<$ 0.5 ml/kg/h for 6 consecutive hours. The onset time is the earliest timestamp meeting any criterion. To strictly prevent label leakage, we implemented a dynamic truncation protocol for positive patients. For a specific onset timestamp $t$, the observation window is truncated at $t - W$, where the prognostic lead time $W \in \{0, 6, 12, 18, 24\}$ hours. This ensures no post-onset physiological artifacts are artificially exposed to the model\cite{lauritsen2020explainable}.

Based on clinical consensus \cite{tomasev2019}, we selected 50 core clinical variables spanning vital signs, laboratory tests, and fluid records (e.g., creatinine, BUN, MAP). Artifacts and outliers were filtered using expert-defined physiological boundaries and treated as missing values.

\begin{figure*}[t] 
    \centering
    \includegraphics[width=0.95\textwidth]{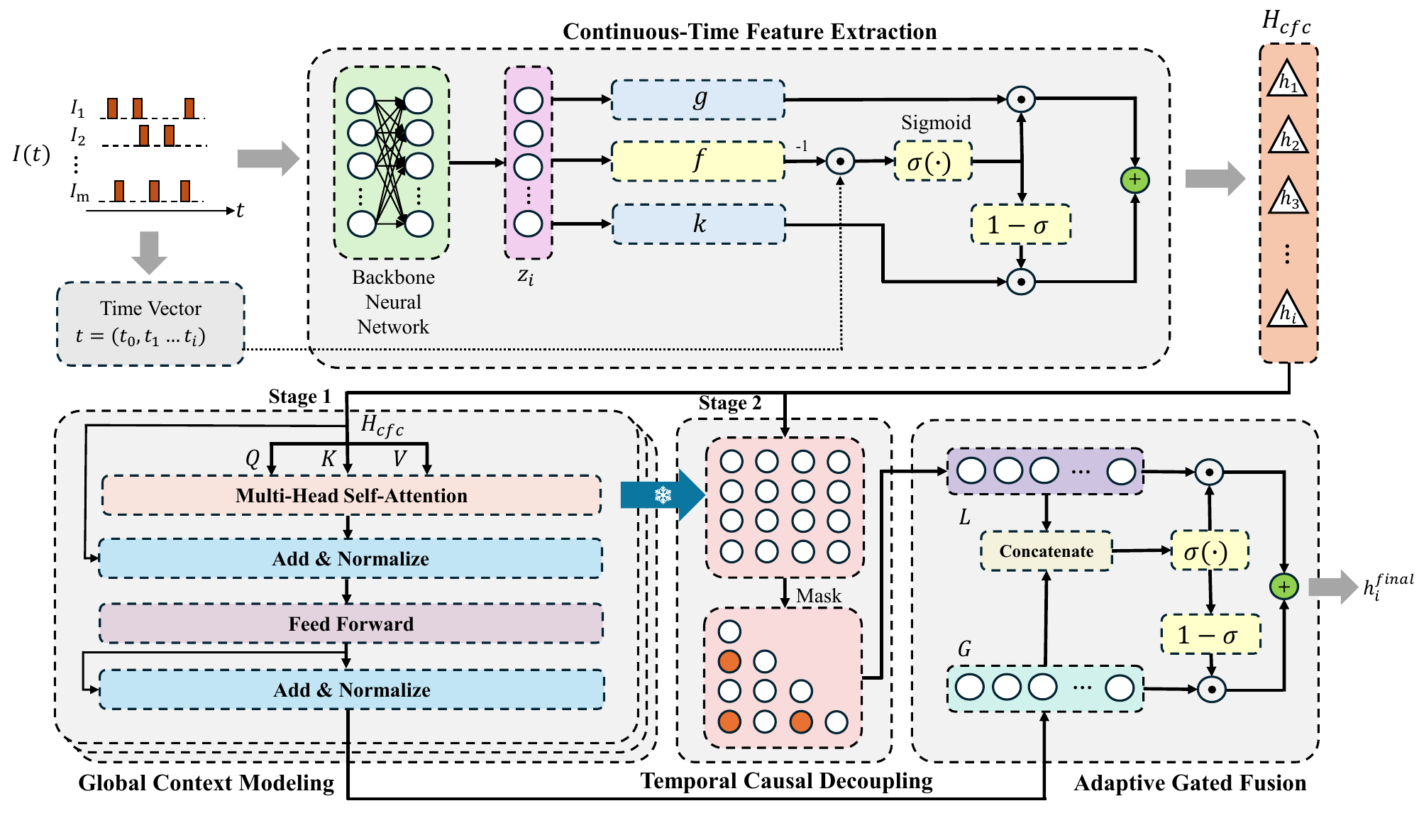}
    \caption{The overall architecture of the Two-Stage Continuous-time Causal-Transformer. Irregular clinical inputs are first natively encoded by the closed-form continuous-time module into continuous physiological states. In Stage 1, the Transformer processes these states to model the patient's global historical trajectory. In Stage 2, a Temporal Causal Decoupling module utilizes a strictly directed mask to generate a causal matrix, explicitly isolating the exact acute physiological triggers to resolve black-box opacity. Finally, an Adaptive Gated Fusion mechanism dynamically balances this transparent causal signal with the global background progression, yielding the final interpretable predictive representation.}
    \label{fig:architecture}
\end{figure*}

Unlike standard discrete models that rely on artificial imputation (e.g., forward-filling) which can distort physiological \cite{lipton2016}, our CT-Former natively processes irregular observations. Continuous features are first Z-score normalized and zero-padded. Instead of utilizing these zeros directly, inputs are governed by an observational mask $\mathbf{M}$, dictating valid measurements for hidden state updates, and a time-delta matrix $\Delta \mathbf{T}$, recording the elapsed time since the last valid observation. By continuously ingesting $\Delta \mathbf{T}$, the framework autonomously manages physiological decay during unobserved intervals, completely eliminating the need for synthetic imputation\cite{che2018recurrent}.

\section{Methodology}
\label{sec:methodology}

To effectively predict Acute Kidney Injury (AKI) from irregularly sampled time series and resolve the black-box opacity of deep learning models, we propose a Two-Stage Continuous-time Causal-Transformer framework, as illustrated in Fig. \ref{fig:architecture}. We first natively encode continuous-time physiological states via CfC and capture global historical trajectories using a Transformer. To overcome the opaque prediction mechanisms, we design a temporal causal decoupling module that generates a directed causal matrix to explicitly trace and isolate the exact acute physiological triggers of AKI\cite{zhangqiyi}. An adaptive gated mechanism then dynamically fuses this transparent causal signal with the broader global trajectory. Finally, we employ a decoupled two-stage training paradigm to guarantee that the interpretable causal structure is independently optimized and rigorously protected from global representation interference.

\subsection{Continuous-Time Feature Extraction}
\label{subsec:cfc}
Standard recurrent neural networks update their hidden states in discrete steps, which may overlook the continuous physiological evolution of patients between observations \cite{lipton2016}. To address this, we adopt the Closed-form Continuous-time (CfC) neural networks, an advanced architecture evolved from the Liquid Time-Constant (LTC) framework \cite{hasani2021liquid}. 

The foundational dynamics of the LTC framework model the hidden state $h(t)$ using an ordinary differential equation (ODE) \cite{chen2018neural}:
\begin{align}
    \frac{dh(t)}{dt} = -\left[ \mathbf{w} + f(x(t), h(t), \theta) \right] \odot h(t) + \mathbf{w} \odot f(x(t), h(t), \theta),
    \label{eq:ltc_ode}
\end{align}
where $\odot$ denotes the Hadamard product, $\mathbf{w}$ is a bias vector parameter, and $f(\cdot)$ is a neural network parameterized by $\theta$. While the LTC ODE intrinsically adapts its decay rate based on incoming stimuli $x(t)$ (mimicking biological synaptic transmission), solving it requires numerical ODE solvers, which can be computationally stiff.

To enhance efficiency, our model directly utilizes the explicit closed-form approximation of these dynamics \cite{hasani2022}. The CfC solution for the hidden state at time step $i$, given the previous state $h_{i-1}^{cfc}$ and the elapsed time $\Delta t_i$, is formulated as:
\begin{align}
    h_i^{cfc} &= \sigma \left( -f(z_i) \odot \Delta t_i \right) \odot g(z_i) \nonumber \\
    &\quad + \left[ 1 - \sigma \left( -f(z_i) \odot \Delta t_i \right) \right] \odot k(z_i),
    \label{eq:cfc_closed_form}
\end{align}
where $z_i =[x_i; h_{i-1}^{cfc}]$ is the concatenation of the current input and the previous hidden state. The functions $f$, $g$, and $k$ are learnable neural network heads sharing a common backbone, and $\sigma$ represents the sigmoid activation function. This closed-form mechanism allows the CfC module to naturally encode irregular time intervals into a robust continuous latent representation sequence $H_{cfc} =[h_1^{cfc}, \dots, h_T^{cfc}]$.

\begin{algorithm}[!t]
\caption{Stage 1: Base Representation Learning}
\label{alg:stage1}
\begin{algorithmic}[1]
\renewcommand{\algorithmicrequire}{\textbf{Input:}}
\renewcommand{\algorithmicensure}{\textbf{Output:}}
\REQUIRE Clinical sequence $X$, Labels $y$
\ENSURE Pre-trained parameters $\Theta_1$, Global context $G$, CfC states $H_{cfc}$, Attention $A$
\STATE // Step 1: Pre-train the base model
\WHILE{not converged}
    \STATE Sample mini-batch $(X, y)$
    \STATE $H_{cfc} \leftarrow \text{CfC\_Encoder}(X)$
    \STATE $H_{trans}, A \leftarrow \text{Transformer}(H_{cfc})$
    \STATE // Predict using the final transformer state
    \STATE $\hat{y} \leftarrow \text{Classifier}(H_{trans, T})$
    \STATE Update $\Theta_1$ by minimizing $\mathcal{L}_{CE}(\hat{y}, y)$
\ENDWHILE
\STATE // Step 2: Extract representations for Stage 2
\FOR{each patient in dataset}
    \STATE Extract continuous-time hidden states: $H_{cfc}$
    \STATE Extract global history context: $G = H_{trans, T}$
    \STATE Extract final layer self-attention matrix: $A$
    \STATE Save tuple $(H_{cfc}, G, A, y)$
\ENDFOR
\end{algorithmic}
\end{algorithm}

\subsection{Global Context Modeling and Self-Attention}
\label{subsec:transformer}
While the CfC module natively captures local continuous dynamics, we introduce a Transformer encoder \cite{vaswani2017attention} to construct a comprehensive global clinical baseline. 

Since the hidden states $H_{cfc}$ are already imbued with exact time-decay information, they are directly processed using Multi-Head Self-Attention bounded by a standard masking matrix $M_{causal} \in \mathbb{R}^{T \times T}$ to prevent future information leakage. The Transformer blocks output the integrated sequence $H_{trans}$, and we extract the representation at the final time step $T$ as our overarching global context vector, denoted as $G = H_{trans, T}$. 

Crucially, while this standard self-attention mechanism excels at mapping holistic sequences, the resulting raw attention matrix $A \in \mathbb{R}^{T \times T}$ merely reflects opaque, pair-wise statistical co-occurrences. This inherently uninterpretable black-box representation lacks the explicit directed reasoning necessary for clinical trust, which directly motivates our structural causal transformation in Stage 2\cite{zichun}.

\subsection{Temporal Causal Decoupling}
The pure attention matrix $A$ relies merely on statistical co-occurrences and lacks the temporal directionality necessary for reasoning. To uncover the explicitly triggering physiological shocks, we introduce a causal discovery mechanism in Stage 2.

\subsubsection{Directed Causal Matrix Generation}
Since causation strictly follows the arrow of time, future events cannot influence past states. We apply a linear transformation weight $W_c$ to purify the statistical attention matrix $A$, bounded by a strictly lower-triangular structural mask $M \in \{0, 1\}^{T \times T}$ (where $M_{i,j}=1$ iff $i>j$). The directed temporal causal matrix $B$ is formulated as:
\begin{equation}
    B = \text{ReLU}(A W_c) \odot M.
\end{equation}
In matrix $B$, the element $B_{i, j}$ represents the directed causal impact of the $j$-th time step on the $i$-th time step.

\subsubsection{Impact Score and Local Causal Vector}
To identify the most critical time step responsible for functional deterioration, we compute the outgoing causal impact score for each step $j$. We aggregate the direct causal influences by summing the columns of the temporal causal matrix $B$:
\begin{equation}
    S_j = \sum_{i=1}^{T} B_{i, j},
\end{equation}
where $S_j \in \mathbb{R}$ quantifies the total downstream impact of the clinical event at time $t_j$. We normalize these impact scores into an attention distribution $\alpha$ using the Softmax function:
\begin{equation}
    \alpha_j = \frac{\exp(S_j)}{\sum_{k=1}^{T} \exp(S_k)}.
\end{equation}
The local focal vector $L \in \mathbb{R}^{d_h}$, which explicitly encapsulates the specific acute mutation event, is then derived by the expectation over the continuous CfC hidden states:
\begin{equation}
    L = \sum_{j=1}^{T} \alpha_j h_j^{cfc}.
\end{equation}

\begin{algorithm}[!t]
\caption{Stage 2: Causal Decoupling and Gated Fusion Training}
\label{alg:stage2}
\begin{algorithmic}[1]
\renewcommand{\algorithmicrequire}{\textbf{Input:}}
\renewcommand{\algorithmicensure}{\textbf{Output:}}
\REQUIRE Extracted tuples $(H_{cfc}, G, A, y)$, Structural Mask $M$
\ENSURE Fusion parameters $\Theta_2$, Causal generation weight $W_c$
\STATE Initialize causal generation network $W_c$, gating network $W_g$, and classifier $W_{cls}$
\WHILE{not converged}
    \STATE Sample mini-batch $(H_{cfc}, G, A, y)$
    
    \STATE // Generate directed Temporal Causal Matrix
    \STATE $B = \text{ReLU}(A W_c) \odot M$
    
    \STATE // Extract local acute causal vector
    \STATE Calculate impact scores $S_j = \sum_{i} B_{i,j}$
    \STATE Normalize attention weights $\alpha_j = \text{softmax}(S_j)$
    \STATE $L = \sum_{j} \alpha_j h_{j}^{cfc}$
    
    \STATE // Adaptive Gated Fusion
    \STATE $g = \sigma(W_g [G; L] + b_g)$
    \STATE $h_{final} = g \odot G + (1 - g) \odot L$
    
    \STATE // Final Prediction and Loss with L1 Penalty
    \STATE $\hat{y} = \sigma(W_{cls} h_{final} + b_{cls})$
    \STATE $\mathcal{L}_{Stage2} = \mathcal{L}_{CE}(\hat{y}, y) + \lambda \|B\|_1$
    \STATE Update $\Theta_2$ and $W_c$ by minimizing $\mathcal{L}_{Stage2}$
\ENDWHILE
\end{algorithmic}
\end{algorithm}

\subsection{Adaptive Gated Fusion and Optimization}
\label{subsec:fusion}
To formulate the final predictive diagnosis without diluting the identified physiological mutation, the isolated local critical state $L$ is dynamically integrated with the terminal global context $G$ using a feature-wise gating mechanism.

We compute a fusion coefficient gate $g \in (0,1)^{d}$ to adaptively balance the local acute signals and broad historical trends:
\begin{equation}
    g = \sigma \left( W_g [G; L] + b_g \right),
\end{equation}
where $W_g$ and $b_g$ are learnable parameters, and $[ \cdot ; \cdot ]$ denotes vector concatenation. The final risk representation $h_{final}$ is derived via an element-wise integration:
\begin{equation}
    h_{final} = g \odot G + (1 - g) \odot L.
\end{equation}

During acute physiological decompensation, specific dimensions of the gate $g$ may adjust to heavily prioritize the transient dynamics encoded within $L$. Conversely, for gradual chronic progression, the gating mechanism seamlessly reverts reliance onto the global trajectory $G$.

The comprehensive predictive probability $\hat{y}$ is obtained by projecting the fused final state through a linear classification head:
\begin{equation}
    \hat{y} = \sigma(W_{cls} h_{final} + b_{cls}).
\end{equation}

During Stage 2, with the base models fully frozen, we exclusively train the causal generation weights $W_c$, the gating network $W_g$, and the classification head. To encourage a sparse and interpretable causal graph where only the most defining critical events emerge, we apply an $L_1$ regularization penalty on the causal matrix $B$. The final loss function for Stage 2 is defined as:
\begin{equation}
    \mathcal{L}_{Stage2} = \mathcal{L}_{CE}(\hat{y}, y) + \lambda \|B\|_1,
\end{equation}
where $\lambda$ controls the structural sparsity, forcing the model to confidently select the minimal set of root causes behind the AKI occurrence.

\begin{table*}[!t]
  \caption{The experimental results involved comparative tests across ML, DL}
  \label{tab:main_results}
  \centering
  \setlength{\tabcolsep}{4pt}
  \renewcommand{\arraystretch}{1.25}
  \begin{tabular}{llcccccccccc}
    \toprule
    & \multirow{2}{*}{Models} & \multicolumn{2}{c}{0h} & \multicolumn{2}{c}{\textit{6h}} & \multicolumn{2}{c}{12h} & \multicolumn{2}{c}{18h} & \multicolumn{2}{c}{24h} \\
    \cmidrule(lr){3-4} \cmidrule(lr){5-6} \cmidrule(lr){7-8} \cmidrule(lr){9-10} \cmidrule(lr){11-12}
    & & AUROC & AUPRC & AUROC & AUPRC & AUROC & AUPRC & AUROC & AUPRC & AUROC & AUPRC \\
    \midrule
    \multirow{2}{*}{ML} & Random Forest \cite{breiman2001random}
    & 0.7538 & 0.5773 & 0.7218 & 0.5307 & 0.7075 & 0.5042 & 0.6949 & 0.4808 & 0.6792 & 0.4414 \\
    & XGBoost \cite{chen2016xgboost}
    & 0.8071 & 0.6303 & 0.7549 & 0.5716 & 0.7370 & 0.5435 & 0.7248 & 0.5159 & 0.7070 & 0.4766 \\
    \midrule
    \multirow{7}{*}{DL} & LSTM \cite{hochreiter1997long}
    & 0.7136 & 0.4723 & 0.7058 & 0.4887 & 0.6806 & 0.4401 & 0.6790 & 0.4323 & 0.6414 & 0.3845 \\
    & ODE-RNN \cite{rubanova2019latent}
    & 0.7765 & 0.5981 & 0.7441 & 0.5347 & 0.7254 & 0.5209 & 0.7115 & 0.4983 & 0.6824 & 0.4682 \\
    & Transformer \cite{vaswani2017attention}
    & 0.7996 & 0.6192 & 0.7532 & 0.5422 & 0.7311 & 0.5316 & 0.7306 & 0.5052 & 0.7043 & 0.4829 \\
    & RKN-$\Delta t$ \cite{becker2019recurrent}
    & 0.8194 & 0.6619 & 0.7948 & 0.5799 & 0.7637 & 0.5862 & 0.7497 & 0.5399 & 0.7208 & 0.5035 \\
    & LTC \cite{hasani2021liquid}
    & 0.8195 & 0.6776 & 0.7871 & 0.6277 & 0.7623 & 0.5921 & 0.7519 & 0.5375 & 0.7186 & 0.5051 \\
    & CfC \cite{hasani2022}
    & 0.8249 & 0.6781 & 0.7892 & 0.6300 & 0.7549 & 0.5811 & 0.7482 & 0.5488 & 0.7235 & 0.5024 \\
    & t-PatchGNN \cite{patchgnn}
    & 0.8538 & 0.7221 & 0.8183 & 0.6648 & 0.7762 & 0.5527 & 0.7658 & 0.5494 & 0.7205 & 0.5070 \\
    \midrule
    & Ours
    & \textbf{0.8872} & \textbf{0.7673} & \textbf{0.8471} & \textbf{0.7073} & \textbf{0.8098} & \textbf{0.6553} & \textbf{0.7782} & \textbf{0.6044} & \textbf{0.7648} & \textbf{0.5858} \\
    \bottomrule
  \end{tabular}
\end{table*}

\section{Experiment}
\label{sec:experiment}
\subsection{Evaluation}
The early prediction of AKI is formulated as a binary classification task, for which the pivotal metric is the Area Under the Receiver Operating Characteristic Curve (AUROC) \cite{fawcett2006introduction}. This metric evaluates the model's discriminative capability across all possible classification thresholds, where a higher AUROC value indicates superior performance in distinguishing between AKI and non-AKI cases. However, given the inherent class imbalance in the MIMIC-IV dataset—where positive AKI cases constitute a strict minority—AUROC alone can present an overly optimistic assessment. Therefore, the Area Under the Precision-Recall Curve (AUPRC) serves as the second critical performance indicator. AUPRC is highly sensitive to the model's ability to accurately capture the minority class without generating excessive false positives, making it the most stringent and clinically relevant metric for imbalanced medical diagnosis tasks \cite{saito2015precision, davis2006relationship}. In the subsequent sections, we present specific experimental outcomes using these two metrics, including a series of comparative analyses against state-of-the-art baselines and ablation studies to substantiate the effectiveness of the proposed CT-Former architecture.

\begin{figure}[htbp]
  \centering
  \subfigure[ROC Curves]{
    \includegraphics[width=0.7\linewidth]{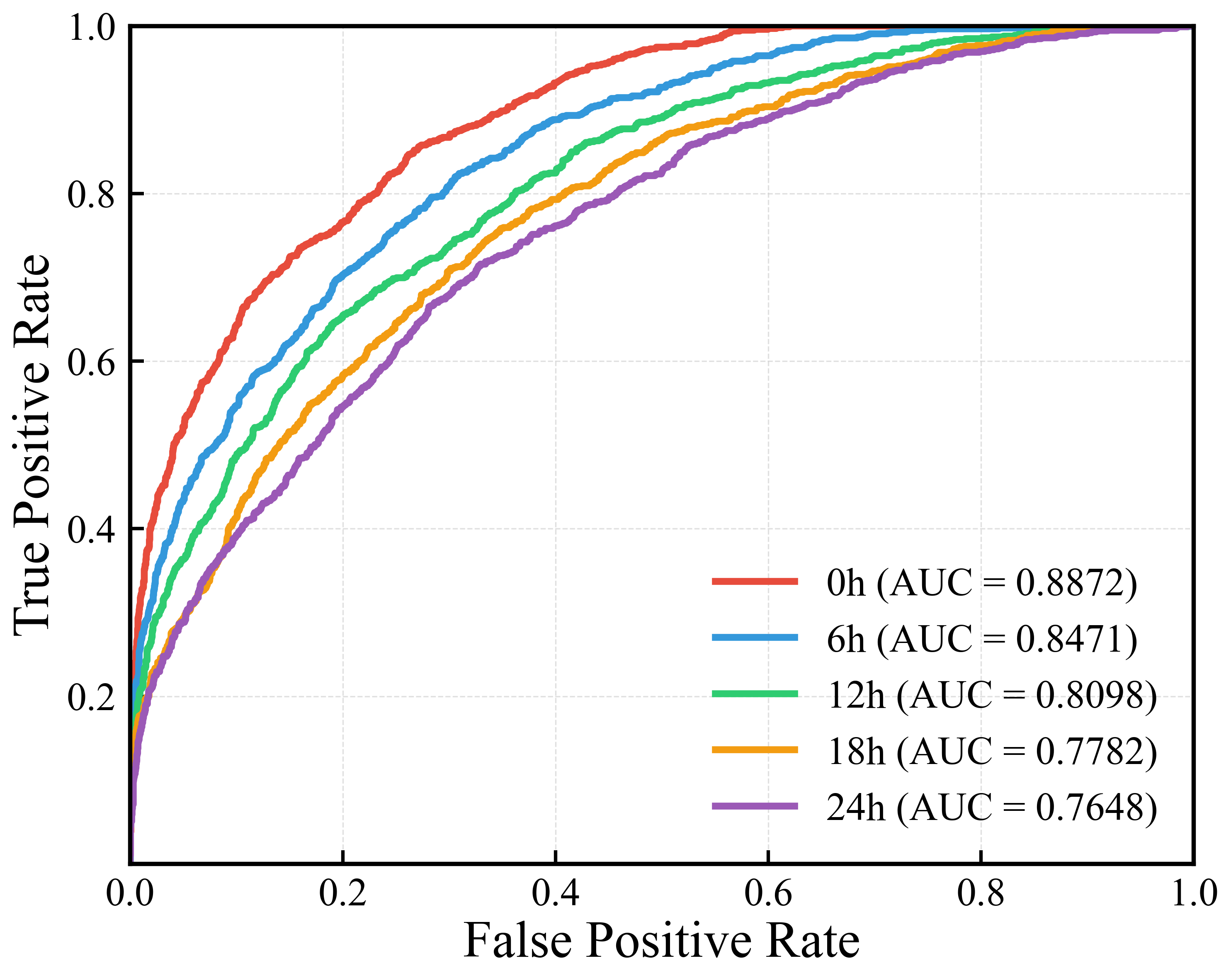}
    \label{subfig:roc_all}
  }
  \vspace{0.2cm}
  \subfigure[Precision-Recall Curves]{
    \includegraphics[width=0.7\linewidth]{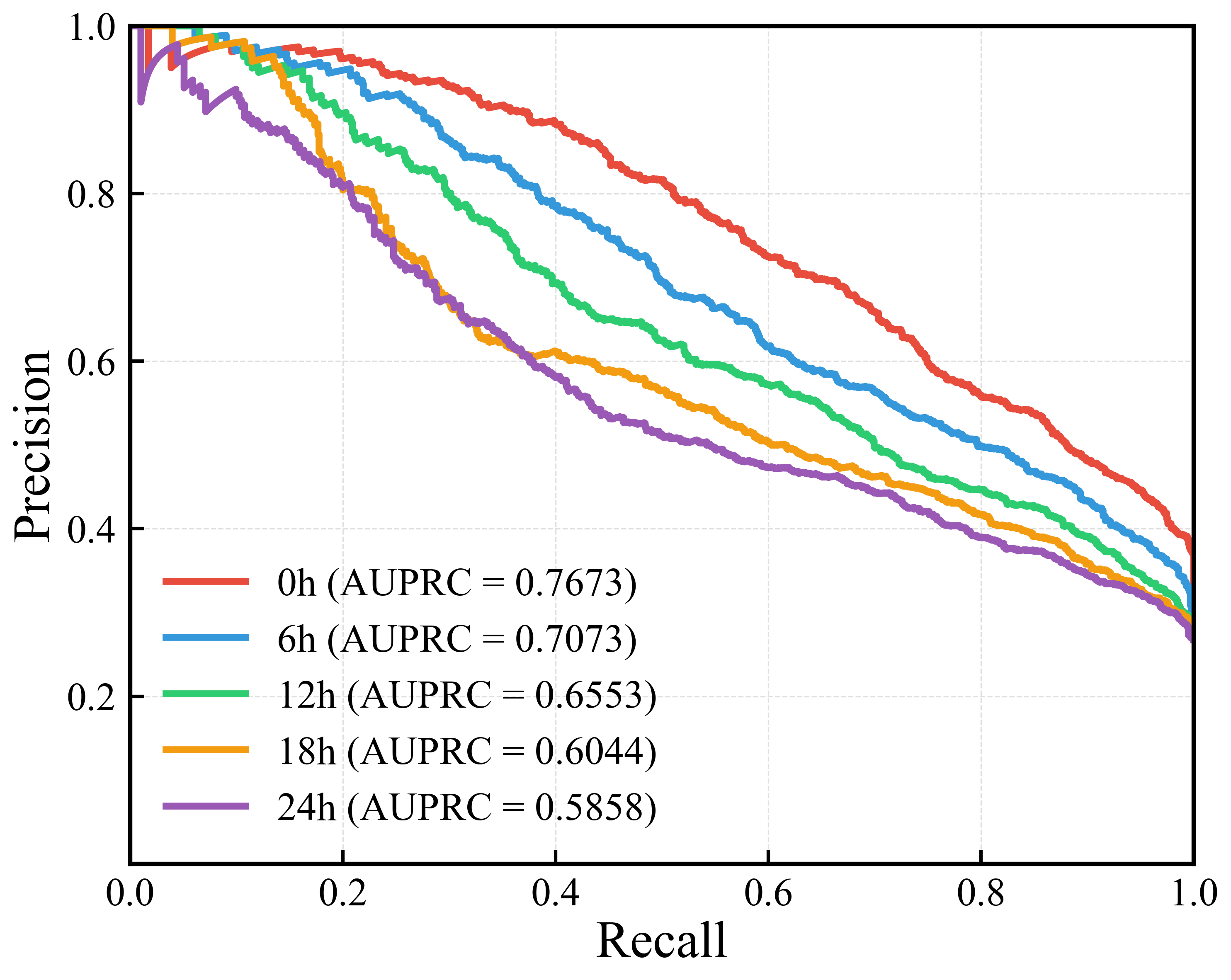}
    \label{subfig:prc_all}
  }
  \caption{Predictive Performance of CT-Former. (a) ROC curves validate consistent early warning capability across five prediction windows. (b) PR curves demonstrate robust identification of the minority AKI class, maintaining high precision at extended lead times.}
  \label{fig:roc_pr_all}
\end{figure}

\subsection{Results}
We trained the proposed model on five distinct prediction horizons ranging from 0 to 24 hours prior to the onset of AKI and visualized the classification performance through ROC and Precision-Recall curves, as illustrated in Fig. \ref{fig:roc_pr_all}. In clinical settings, timely diagnosis must be established as early as possible to allow for effective intervention. The ROC curves across the five tasks demonstrate a natural temporal correlation: predictive discriminative power progressively strengthens as the observation window approaches the event onset. Despite the increasing difficulty associated with longer lead times, the curves for the 18-hour and 24-hour windows remain well above the random baseline. Furthermore, given the inherent class imbalance in AKI datasets, the sustained high area under the PR curves across early prediction windows demonstrates that our method maintains strong recognition capability for the clinically critical minority class while minimizing false alarms.

The quantitative results are presented in Table \ref{tab:main_results}. Our model demonstrates substantial improvements over comparative methods evaluated under identical protocols. Among all baselines, the standard LSTM exhibited the weakest performance because its fixed-interval assumption forcibly distorts the sporadic ICU data. Traditional machine learning models presented a mixed profile. For instance, Random Forest achieves moderate performance, while XGBoost achieves relatively strong short-term results but degrades sharply as the prediction window extends to 24 hours due to its inability to capture sequential dependencies. Similarly, the vanilla Transformer demonstrates limited efficacy because its standard discrete positional encodings fail to natively represent continuous physiological time-lapses.

In contrast, continuous-time models explicitly learn time-decay dynamics. By natively incorporating the elapsed time between observations into their hidden state updates, architectures like ODE-RNN, LTC, and standalone CfC effectively mitigate information staleness and consistently outperform standard discrete methods. Meanwhile, RKN-$\Delta t$ and t-PatchGNN emerge as strong competitors by leveraging temporal uncertainty modeling and spatial-temporal graph structures, respectively. Yet, continuous recurrent models eventually suffer from vanishing memory over extended sequences, while patching models exhibit noticeable performance decay at longer prognostic horizons.

Our proposed CT-Former establishes a clear lead over all baselines. This lead spans across every metric and prediction window. The model maintains a superior AUROC of 0.7648 at the challenging 24 hour horizon. This sustained robustness is directly attributed to our novel two stage causal decoupling design. This design systematically resolves the opaque black box limitations of standard end to end continuous architectures. These traditional architectures rely heavily on uninterpretable statistical correlations. The continuous module acts as a highly sensitive encoder. It explicitly tracks irregular physiological fluctuations. Our two-stage paradigm completely abandons purely entangled statistical associations. It instead purifies the global attention field using a strictly directed temporal mask. The network computes the downstream causal impact scores. It then explicitly locks onto the exact acute physiological mutations. An adaptive gate dynamically fuses this isolated local causal signal with the comprehensive global patient trajectory. The CT-Former successfully establishes a transparent causal decision pathway. It simultaneously maintains high sensitivity to extreme clinical events a full day before onset.

\begin{figure}[htbp]
  \centering
  \includegraphics[width=0.7\linewidth]{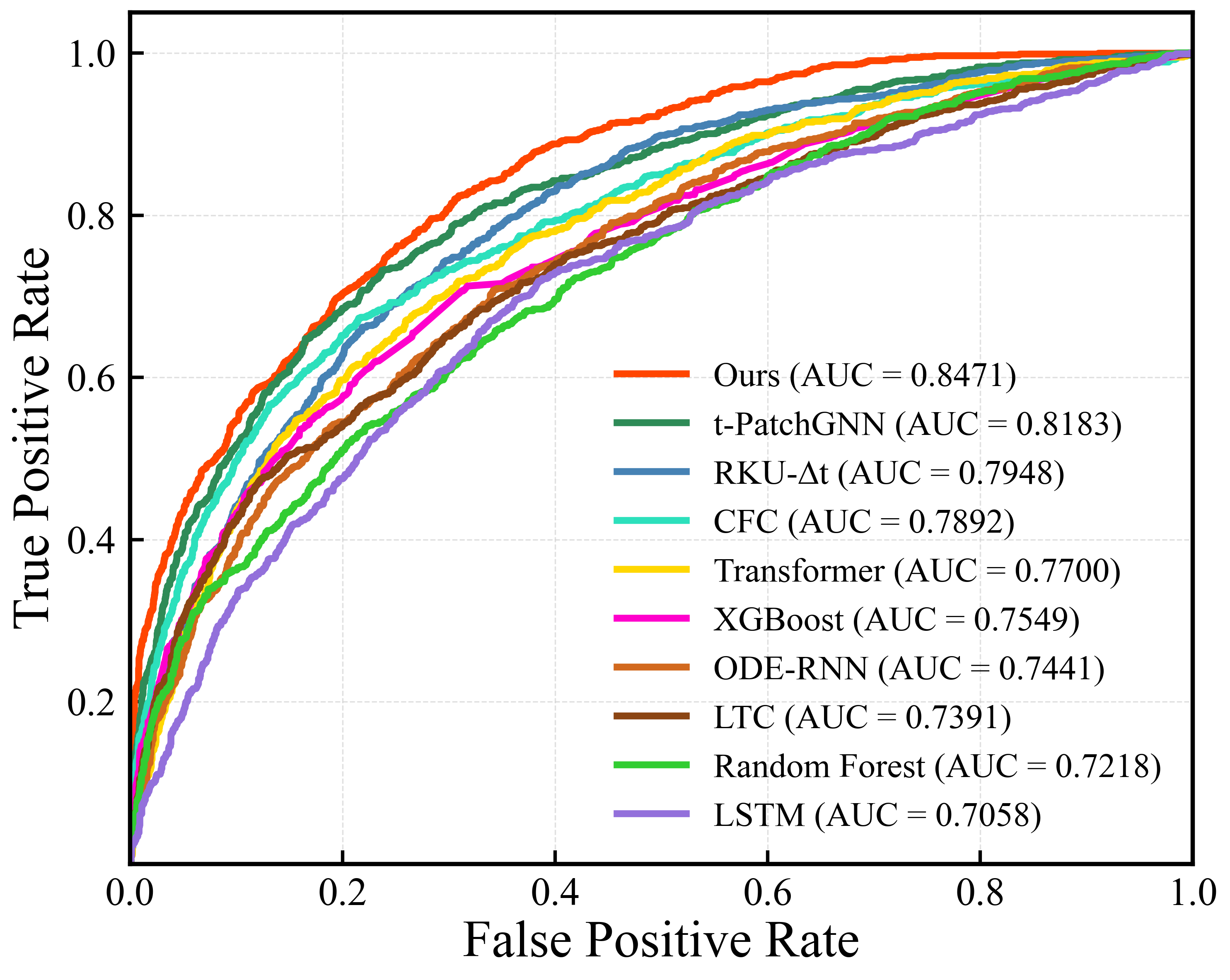}
  \caption{ROC Curve Comparison for the 6-hour Prediction Window. The proposed CT-Former achieves the highest AUROC compared to state-of-the-art baselines. Similar performance advantages were consistently observed across all prediction windows.}
  \label{fig:roc_6h}
\end{figure}

To provide a granular visualization of model behavior, we focus on the ROC curves for the six hour prediction window in Fig. \ref{fig:roc_6h}. The plotted curves illustrate a performance hierarchy defined by the architectural capabilities of the models. The continuous model family forms a closely clustered group by explicitly capturing physiological decays over time, slightly surpassed by the structural t-PatchGNN baseline. Nevertheless, our proposed CT-Former distinctly surpasses all competitors. Visually, the CT-Former curve stands entirely above the others, maintaining a noticeably higher true positive rate at low false positive rates. This steep initial rise is critical for clinical early warning systems because it ensures aggressive sensitivity to at risk patients while severely regularizing false alarms. This visualization provides direct evidence confirming the efficacy of our temporal causal decoupling strategy. By generating a strictly directed causal matrix to explicitly lock onto the exact localized pathological mutation rather than relying on lossy standard sequence poolings, our model prevents the dominant global context from diluting transient shocks. The subsequent adaptive gated fusion captures both sudden acute collapses and gradual chronical deterioration far more reliably.

\vspace{-12pt}
\subsection{Ablation Analysis}

\begin{figure}[htbp]
  \centering
  \subfigure[AUROC Trend]{
    \includegraphics[width=0.46\linewidth]{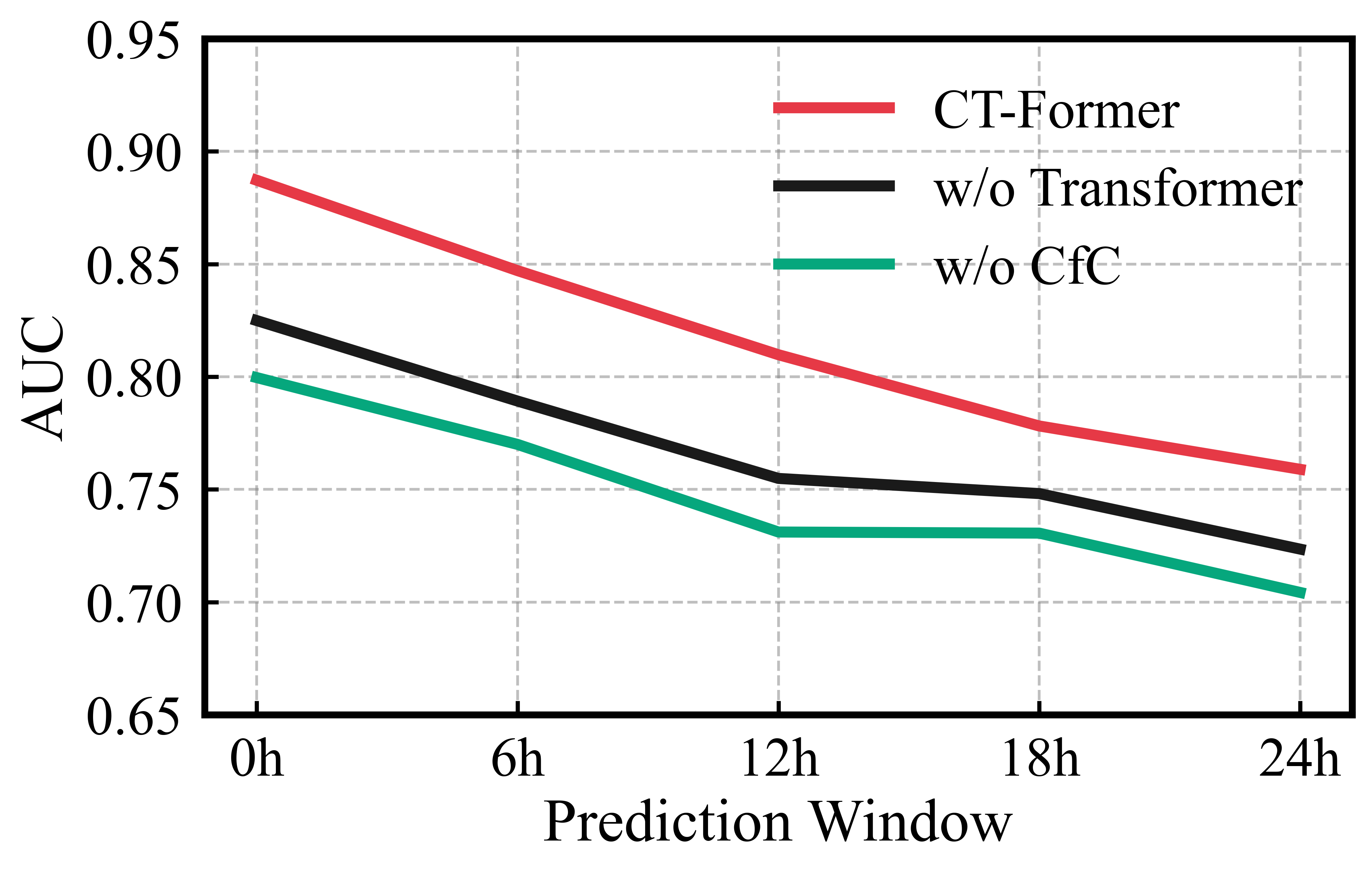}
    \label{fig:ablation_auc}
  }
  \hfill
  \subfigure[AUPRC Trend]{
    \includegraphics[width=0.46\linewidth]{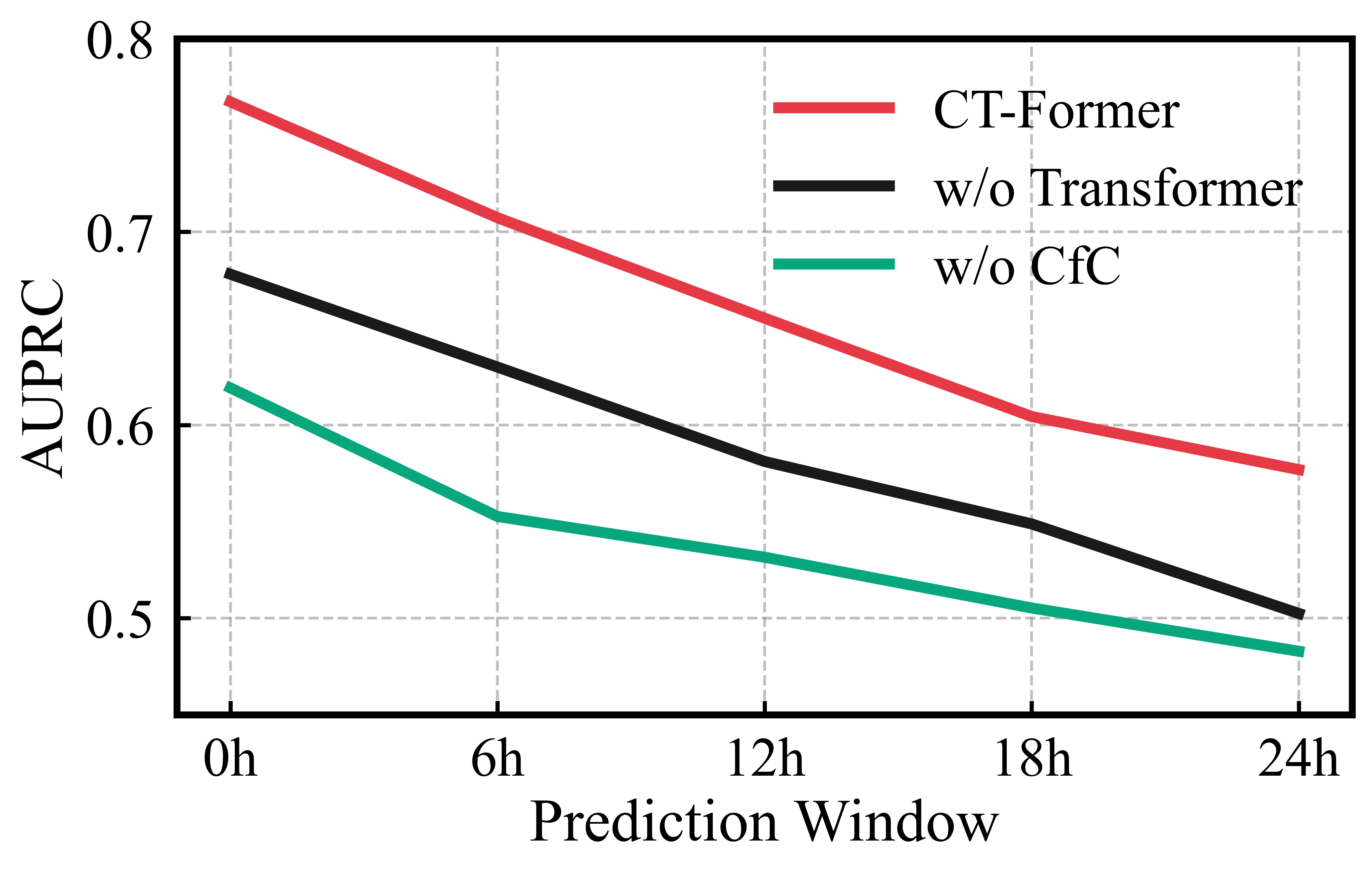}
    \label{fig:ablation_auprc}
  }
  \caption{Ablation Study Results. Performance trajectories of the full CT-Former model compared to variants without the Transformer and without the CfC module across all prediction windows.}
  \label{fig:ablation_trend}
\end{figure}

\textbf{Impact of Model Architecture Components.} To dissect the contribution of each design choice, we conducted an ablation study across all five prediction horizons. We compared the full two stage causal model against variants lacking either the Transformer module or the continuous CfC encoder. As visualized in Fig. \ref{fig:ablation_trend}, removing either core component leads to a severe drop in predictive capability.

The variant excluding the CfC encoder exhibited optimal performance degradation, particularly in the short prediction windows of 0 to 6 hours. Without continuous dynamics explicitly tracked by the CfC, the network treats variable time intervals as uniform steps. It fails to capture the natural decay of physiological signals, losing sensitivity to immediate rapid fluctuations that often trigger acute decompensation.

Conversely, removing the Transformer caused performance to drop notably in the extended windows of 18 to 24 hours. The standalone CfC encoder effectively captures local dynamics but restricts information retention over long sequences. More critically, the absence of the Transformer means the model loses both the global context $G$ and the raw attention matrix $A$. Without $A$, our temporal causal decoding module cannot generate the directed causal matrix $B$. The entire architecture thus loses its ability to compute causal impact scores and lock onto the local acute vector $L$.

\begin{figure}[htbp]
  \centering
  \includegraphics[width=0.95\linewidth]{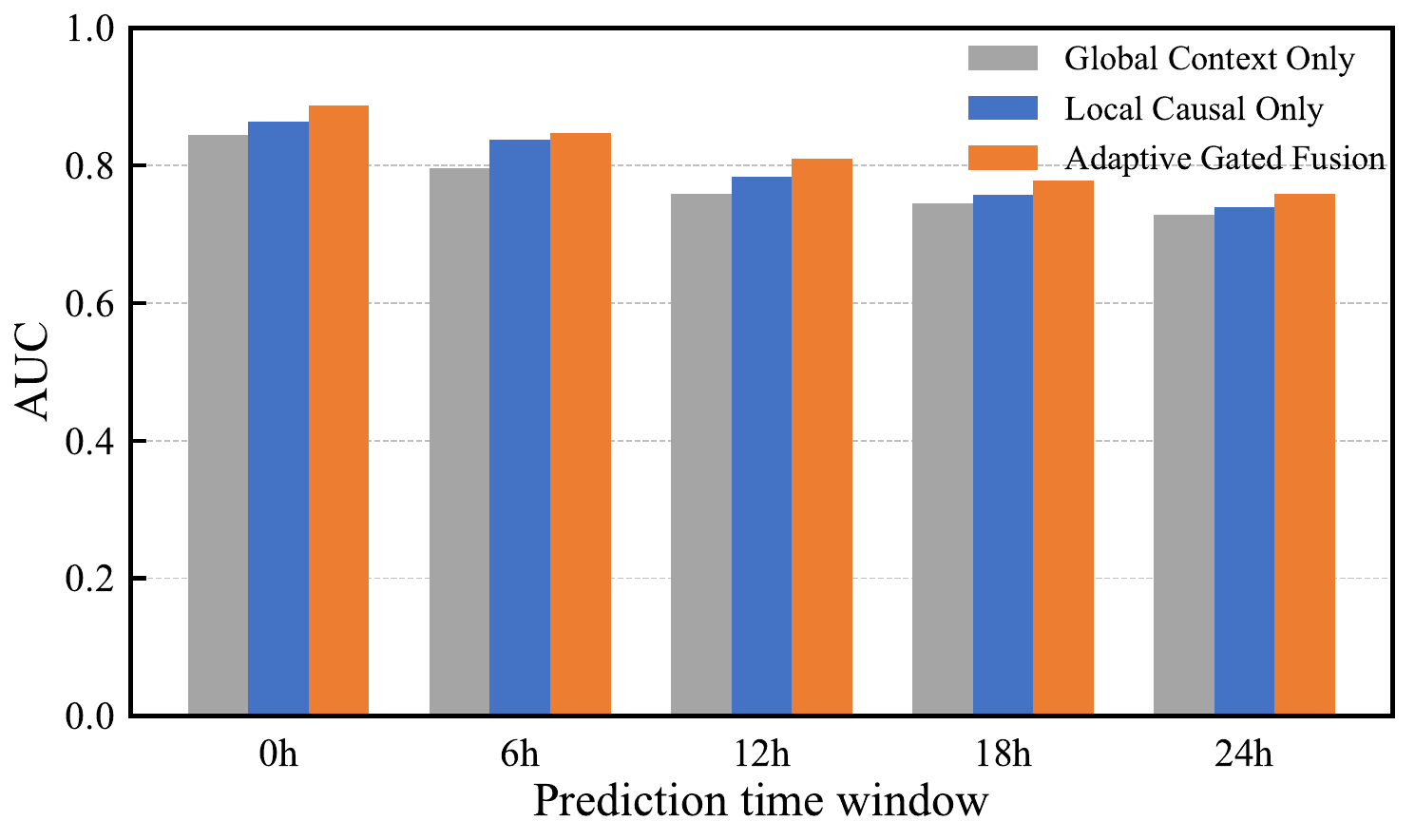}
  \caption{AUC comparison analyzing the causality aware fusion strategy across all prediction windows. The results demonstrate that relying exclusively on the global context causes critical feature dilution and yields the lowest performance. Our adaptive gating mechanism consistently achieves superior AUCs by robustly integrating the locked local causal mutation with the global baseline trajectory.}
  \label{fig:ablation_gating}
\end{figure}

\textbf{Effectiveness of Causality Aware Gated Fusion.} To validate the integration strategy in our second stage, we evaluated the adaptive gating mechanism against two distinct baseline variants: one relying exclusively on the global context $G$, and another relying exclusively on the locked local causal vector $L$. As illustrated in Fig. \ref{fig:ablation_gating}, the architecture incorporating the adaptive gate consistently yields superior AUC scores across all prediction horizons, while the variant relying solely on the global context $G$ performs the worst.

This performance hierarchy is rooted in the temporal heterogeneity of AKI progression. Relying entirely on the global trajectory $G$ averages out the entire sequence, severely diluting sudden physiological shocks and resulting in the lowest predictive sensitivity. Conversely, relying only on the local causal vector $L$ captures acute mutations but suffers from a lack of historical context, potentially raising false alarms for patients with chronically stable abnormal baselines. 

Our adaptive gate resolves this dilemma by acting as a dynamic feature arbitrator. During intense decompensation events, the gate autonomously upweights the localized causal vector $L$ extracted via our directed causal matrix $B$. When insidious baseline shifts dominate the clinical picture, it systematically reverts priority to the holistic global trajectory $G$. By explicitly decoupling the causal shock extraction from the global baseline and adaptively fusing them, our framework gracefully handles diverse pathological pathways without diluting critical warning signals.

\begin{figure}[htbp]
  \centering
  \includegraphics[width=0.9\linewidth]{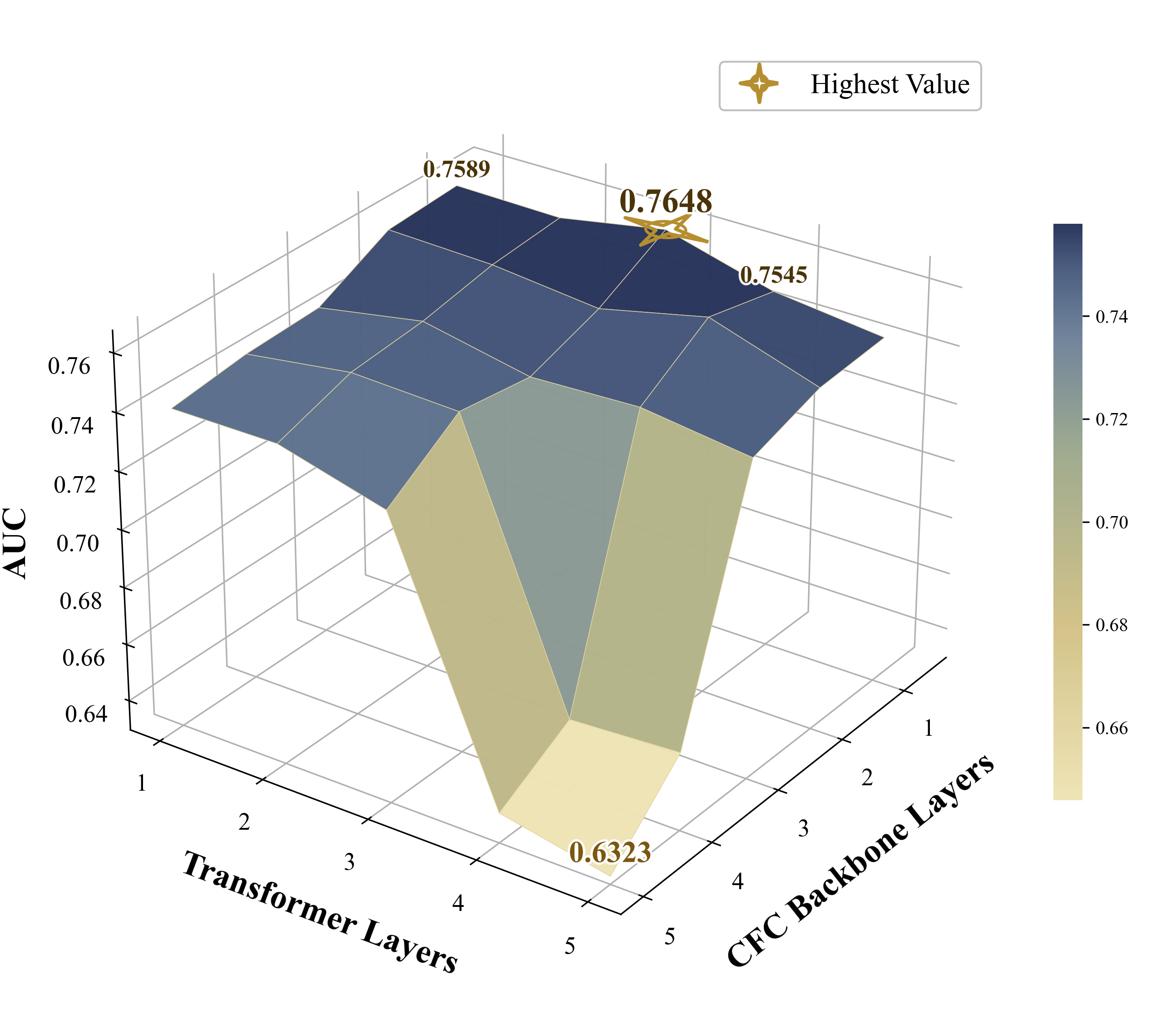}
  \caption{AUROC optimization 3D surface for network depth configurations. We select the highly sensitive 24 hour prediction window as the evaluation background. This architecture comprises one CfC layer and three Transformer layers. Deepening both modules causes severe representation collapse.}
  \label{fig:depth_heatmap}
\end{figure}

\textbf{Network Depth and Representation Capacity.} In continuous medical time-series frameworks, deeper networks do not strictly correlate with better generalization due to numerical stiffness and over-parameterization. We evaluated the interaction between the number of CfC layers and the Transformer layers shown in Fig. \ref{fig:depth_heatmap}. The architecture achieves a peak AUROC of 0.7648 with a highly asymmetrical design utilizing a single-layer continuous backbone paired with a three-layer causal attention module. Conversely, performance collapses to 0.6323 when both modules are deepened to five layers.

A single-layer CfC backbone is exactly optimal for computing elapsed-time gates and absorbing irregular intervals. Increasing the depth of closed-form ODE backbone layers exacerbates internal state transitions, which imposes an artificial smoothing effect that degrades the essential sharpness penalty in the mutation intensity score $s_j$. On the other hand, a three-layer Transformer is crucial for resolving complex conditional dependencies efficiently. This degree of multi-layer cross-referencing is mathematically required to distill a sequence of more than 50 variates over a 24-hour window into a reliable causal graph $\tilde{A}$ and a subsequent interaction matrix $C$. While single-layer attention fails to discard confounding correlations, extending the attention module to five layers causes over-integration and destroys the local gradients required for early event identification. This rigorous balance ensures the mechanism remains sensitive to microscopic physiological crashes while comprehending macroscopic decay.

\subsection{Model Interpretability}
The deployment of deep learning in critical care is frequently impeded by opaque decision-making, engendering a trust deficit among clinicians. While standard feature attribution methods like SHAP \cite{lundberg2017} are widely used, they treat input features as static and independent. This flawed assumption obliterates the continuous temporal dependencies underlying disease progression and causes a computationally intractable combinatorial explosion ($2^{T \times F}$) for long medical sequences \cite{aas2021explaining}.

We addressed these limitations using TimeSHAP \cite{bento2021} for sequential architectures. Its temporal coalition pruning algorithm avoids exact exhaustive computation. It recognizes the diminishing influence of distant physiological events. Older records are thus aggregated into a single background baseline. This effectively focuses computational resources on critical recent fluctuations. TimeSHAP crucially decomposes risk across feature importance, temporal-level trajectories, and specific feature-time contributions. This interpretability rigorously validates the clinical coherence of our architecture.

\begin{figure}[htbp]
    \centering
    \includegraphics[width=0.75\linewidth]{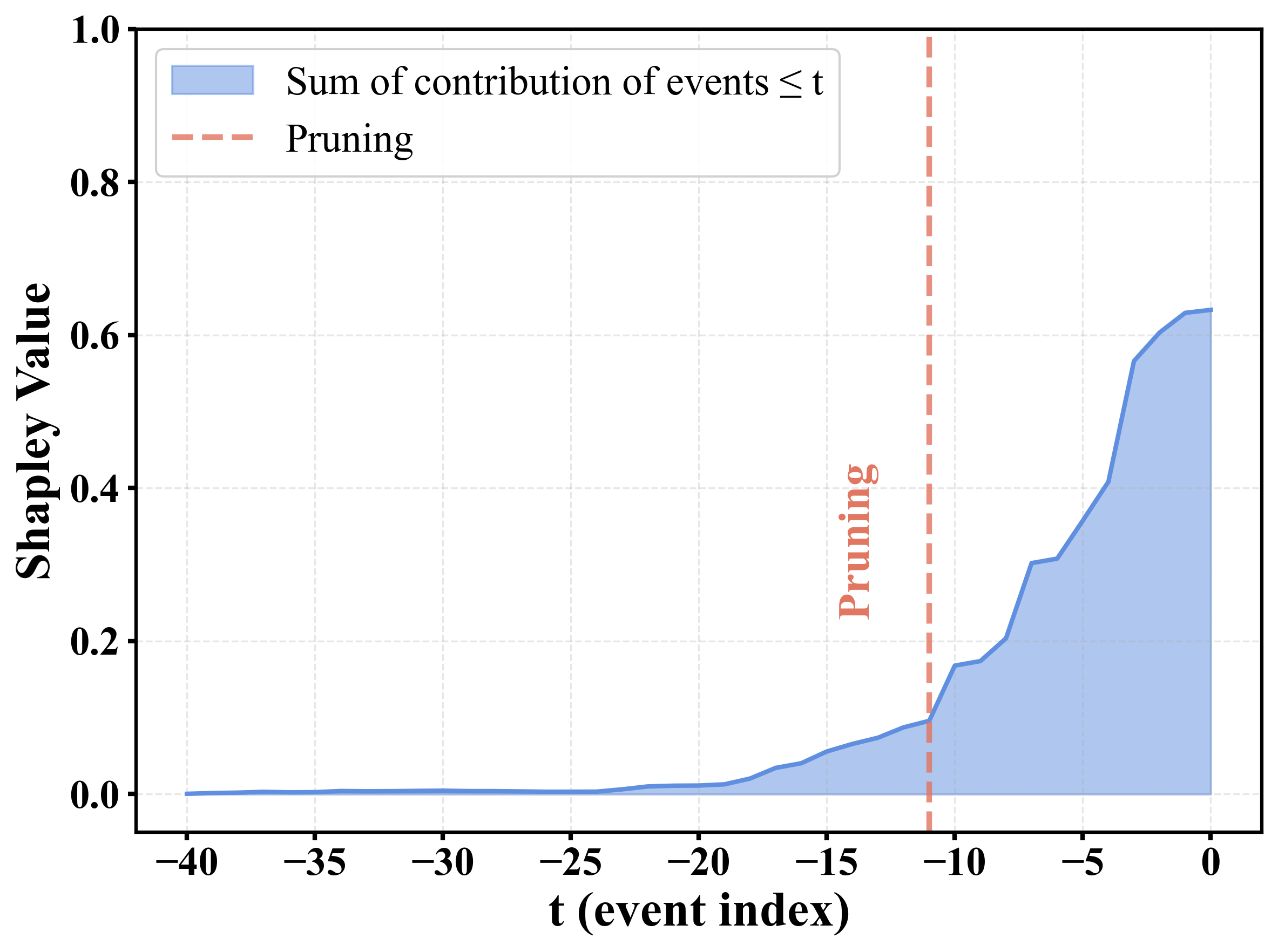}
    \caption{Temporal Pruning Analysis for Patient 30696672. The curve reveals that physiological fluctuations prior to $t=-10$ hours are treated as stable background context, while a sharp accumulation of risk begins at the 11th hour, marking the onset of acute deterioration.}
    \label{fig:shap_pruning}
\end{figure}

To validate the clinical coherence of our model and ensure a rigorous head-to-head comparison between the model's internal architecture and post-hoc explanations, we conducted a granular, three-stage visualization analysis strictly using the temporal records of the exact same patient. Specifically, we selected a representative true-positive AKI case (Patient ID: 30696672), who was assigned a high risk probability of 0.8961.

\subsubsection{Temporal Pruning and Causal Matrix Alignment}
We first examined the temporal scope of the model's decision-making process using the pruning graph shown in Fig. \ref{fig:shap_pruning}. As observed, the cumulative contribution curve of Shapley values remains remarkably flat prior to the timestamp of -10. Consistent with the TimeSHAP pruning theory, this plateau demonstrates that the model explicitly recognizes the distant, stable physiological period as a negligible background context rather than active disease drivers.

The core validation of our causal architecture lies within the subsequent active temporal window. Conventional machine learning and deep learning models operate as opaque statistical black boxes. Their predictions rely on highly entangled mathematical associations. These unconstrained associations inevitably conflate true pathophysiological triggers with spurious correlations. Consequently, their risk trajectories manifest as a continuous statistical blur. In stark contrast, the risk accumulation for CT-Former clearly diverges from this black-box behavior. It exhibits precise non-monotonic escalations at specific isolated intervals. These episodic spikes occur notably at 7 hours and 3 hours prior to prediction.

\begin{figure}[htbp]
    \centering
    \includegraphics[width=0.98\linewidth]{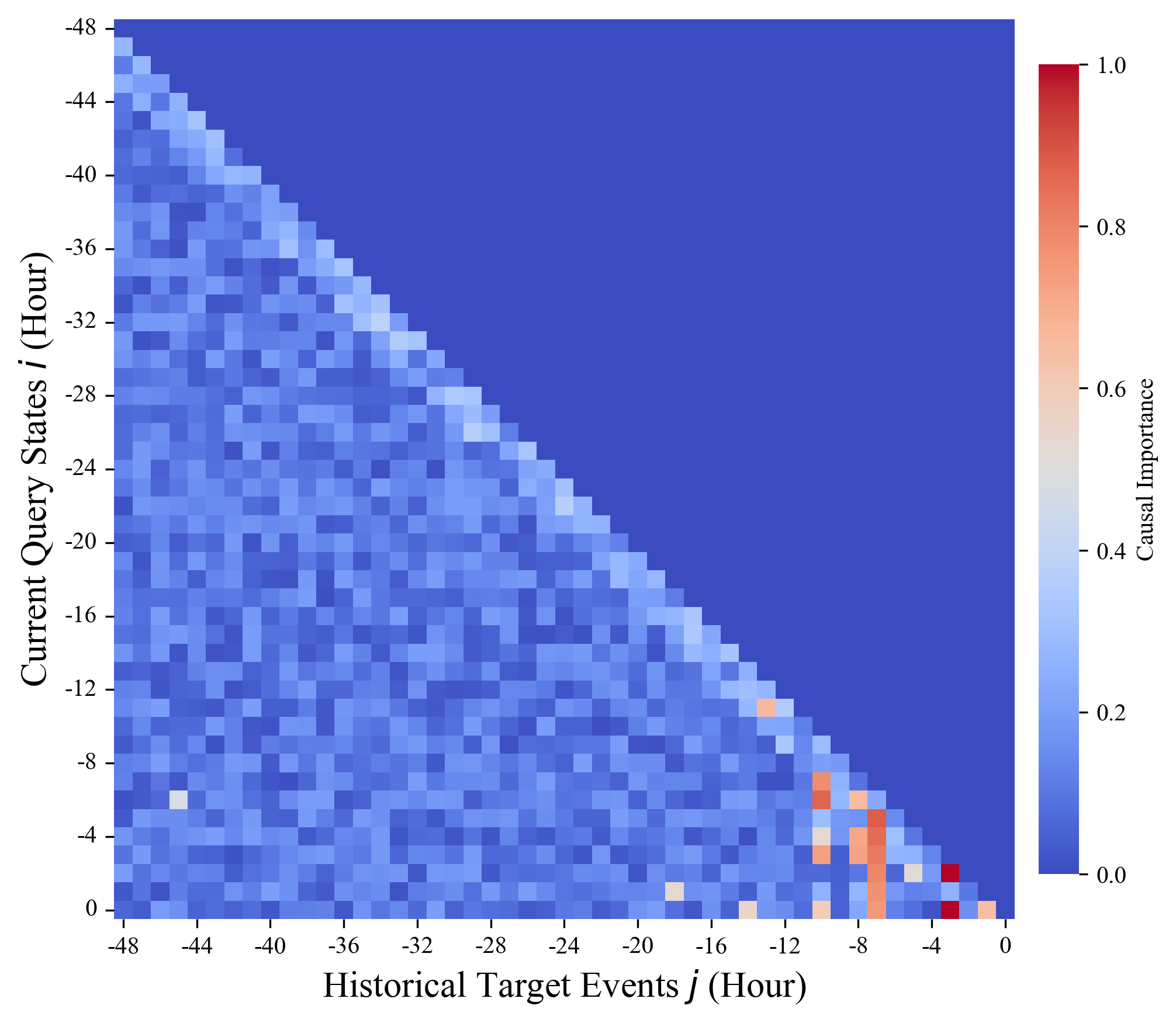} 
    \caption{The autonomously learned Temporal Causal Matrix ($B$). The strictly zero-valued upper-right region enforces temporal precedence. The matrix shows high causal importance along specific vertical columns at $t=-10$, $t=-7$, and $t=-3$, which align with the critical event spikes identified independently by TimeSHAP.}
    \label{fig:causal_matrix}
\end{figure}

This crisp statistical trajectory laterally proves that our model refuses to perform opaque data assimilation. To explicitly verify this, we juxtaposed these post-hoc statistical observations with the model's autonomously learned Temporal Causal Matrix during inference, as visualized in Fig. \ref{fig:causal_matrix}. Rather than hiding decision-making processes within dispersed and uninterpretable neural weights, our Temporal Causal Decoupling module strictly enforces transparent reasoning. The matrix vividly projects intense, concentrated causal weights along distinct vertical columns that correspond flawlessly to the exact anomalous timestamps independently flagged by TimeSHAP.

The perfect structural alignment between the external statistical explainer and the internal directed matrix rigorously proves our primary methodological motivation. It mathematically confirms that CT-Former successfully shatters the black-box paradigm inherent in standard deep learning—formulating highly transparent, causally explainable predictions by intrinsically locking onto true acute physiological mutations.

\subsubsection{Feature Level Attribution}
\begin{figure}[htbp]
    \centering
    \includegraphics[width=0.98\linewidth]{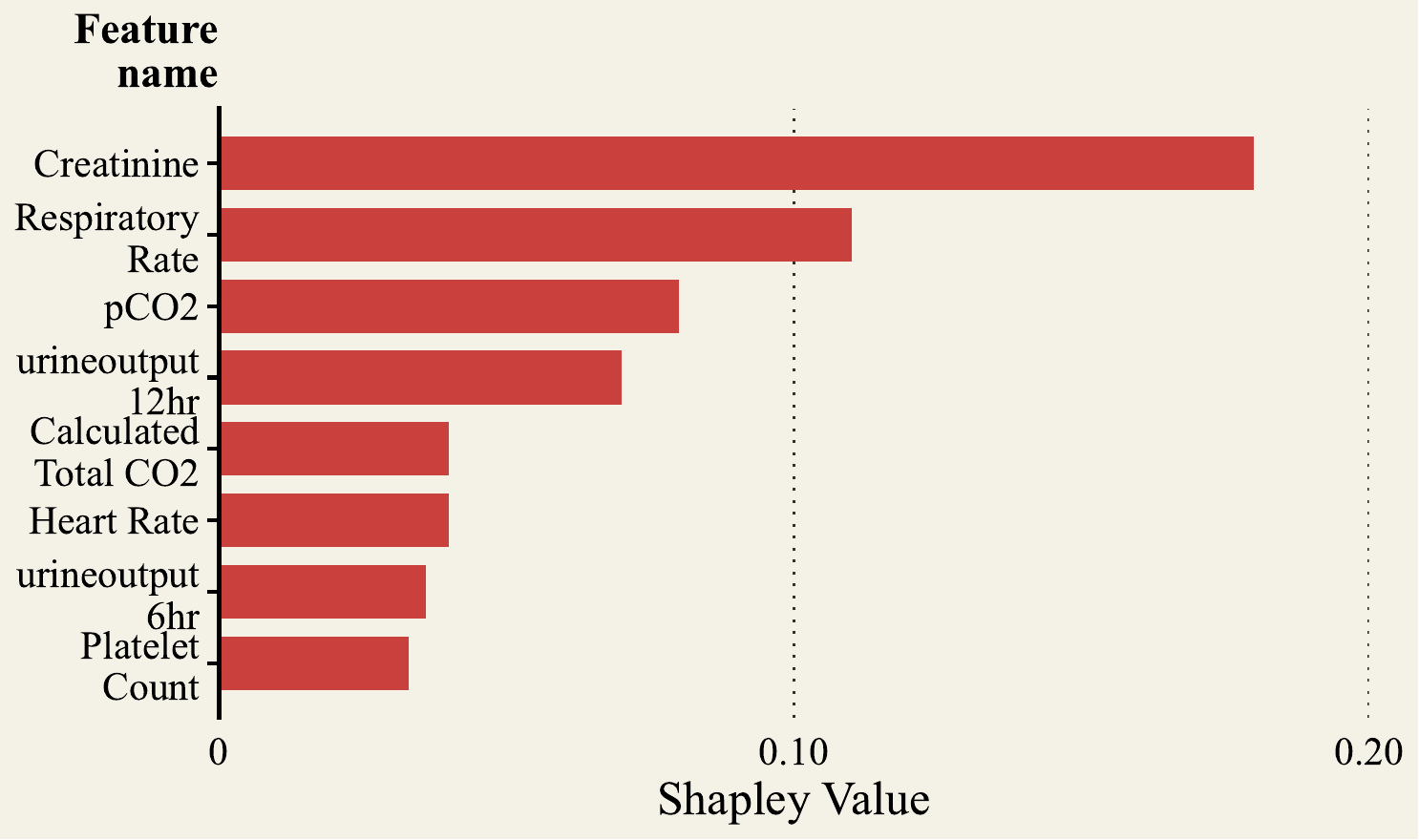}
    \caption{Macroscopic Interpretability. Feature level attribution confirms Serum Creatinine as the dominant predictor. It also identifies respiratory markers linked to metabolic acidosis compensation.}
    \label{fig:shap_feature}
\end{figure}

We further decompose the predictive risk into feature level attributions. This process identifies the exact physiological drivers. The results are visualized in Figure \ref{fig:shap_feature}.

TimeSHAP systematically perturbs a specific clinical variable across all unpruned temporal events simultaneously. It then measures the aggregate impact on the final prediction. This computes the global feature importance. Serum Creatinine emerges as the most dominant predictor with the highest positive Shapley value. The KDIGO clinical practice guidelines diagnose acute kidney injury primarily based on acute fluctuations in serum creatinine. Our model demonstrates a robust alignment with these clinical standards. This alignment establishes the fundamental trustworthiness of the causal architecture.

The model also assigns unexpectedly high importance to Respiratory Rate and partial pressure of Carbon Dioxide. This reveals a sophisticated systemic understanding from a pathophysiological perspective. Severe renal hypoperfusion and dysfunction inevitably impair the excretion of organic acids. This process precipitates metabolic acidosis. The body triggers hyperventilation as an immediate compensatory mechanism for acidosis. This increased respiratory rate expels carbon dioxide and regulates blood pH. Our architecture heavily weighs these respiratory indicators. The network successfully captures the systemic syndrome of acidosis induced respiratory distress. This systemic condition typically precedes overt renal failure. The model proves it is not merely memorizing delayed kidney markers.

\subsubsection{Cell-Level Attribution}
While pruning-level and feature-level attributions isolate critical time points and globally relevant variables, they do not identify which specific feature drove the risk at a given moment. To address this, we compute cell-level attributions, which quantify the marginal contribution of individual clinical variables at precise time steps.

\begin{figure}[htbp]
    \centering
    \includegraphics[width=0.99\linewidth]{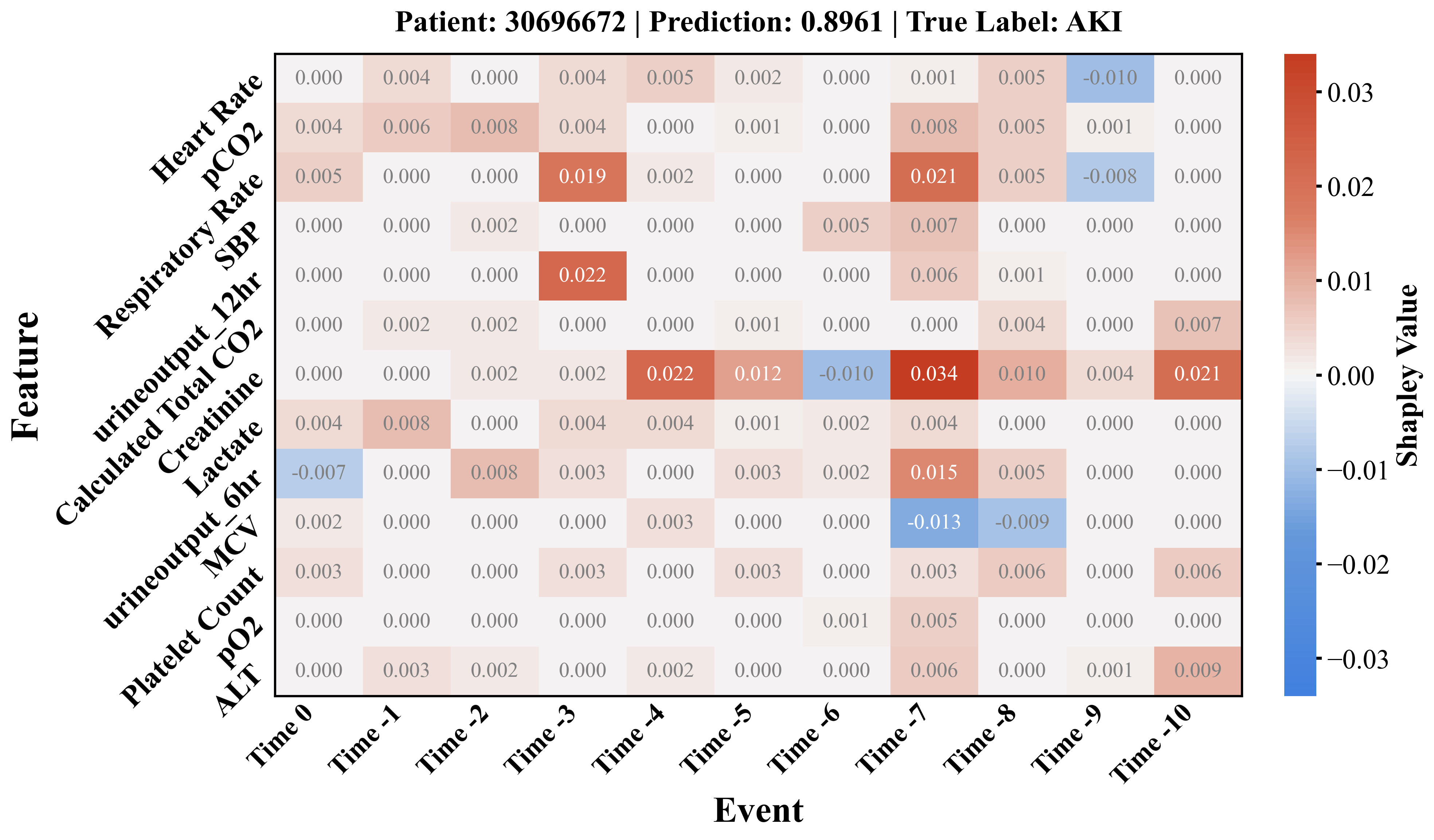}
    \caption{Cell-level attribution for Patient 30696672. The heatmap details specific feature-time risk contributions. High attributions for Serum Creatinine and pCO2 at $t=-7$, followed by Respiratory Rate and urine output at $t=-3$, demonstrate a sequential pathological trajectory of renal impairment and subsequent respiratory compensation.}
    \label{fig:shap_cell}
\end{figure}

The cell-level heatmap (Fig. \ref{fig:shap_cell}) provides a fine-grained analysis of the patient's physiological trajectory. The critical event previously identified at $t=-7$ is predominantly driven by simultaneously high risk attributions from both Serum Creatinine and pCO2. This specific feature-time intersection indicates that the model detects a concurrent deterioration in both renal excretory function and metabolic status.

Subsequently, at $t=-3$, the highest risk attributions shift to Respiratory Rate and 12-hour urine output. From a clinical perspective, this sequence points to a well-defined disease progression: the sustained decrease in urine output reflects continuous declining renal clearance, a primary indicator of AKI. Concurrently, the elevated importance of respiratory rate indicates a systemic compensatory response—specifically, hyperventilation triggered by worsening metabolic acidosis. 

By localizing these specific pathological interactions, the cell-level attribution confirms that CT-Former bases its predictions on coherent, verifiable clinical physiological sequences rather than spurious statistical patterns. This micro-level transparency provides a robust and trustworthy rationale for the model's decision-making process.

\section{Conclusion} \label{sec:conclusion}In this study, we propose the CT-Former architecture. This model integrates continuous time networks with a causal transformer. This combination overcomes the opaque black box nature of standard deep learning in early acute kidney injury prediction. A novel two stage causal decoupling mechanism generates a transparent physiological trajectory. The network explicitly isolates critical acute deteriorations from stable background trends. Extensive validation on the MIMIC-IV cohort demonstrates superior predictive performance. The model consistently outperforms state of the art baselines. The internally learned temporal causal matrix aligns explicitly with independent TimeSHAP evaluations. This structural alignment rigorously validates the interpretability motivation of our study. The transparent architecture successfully captures true clinically underlying causes. It entirely avoids spurious statistical correlations. Future work will integrate multimodal clinical notes. We will also conduct multi center validation. These efforts will advance the model towards a fully interpretable decision support system for critically ill populations.

\bibliographystyle{IEEEtran}
\bibliography{references}
\end{document}